\documentclass[letterpaper, 10 pt, conference]{ieeeconf}
\IEEEoverridecommandlockouts
\let\labelindent\relax
\usepackage{enumitem}
\usepackage{balance}
\usepackage[font={small}]{caption}
\usepackage{subcaption}
\usepackage{array}
\usepackage{textcomp}
\usepackage{mathtools, nccmath}
\usepackage{graphicx}
\usepackage{amsfonts}
\usepackage{amsmath}
\usepackage{amssymb}
\usepackage{algorithm}
\usepackage{algorithmic}
\usepackage{hyperref}
\usepackage{tikz}
\usepackage{arydshln}
\usepackage{multirow}
\usepackage{bm}
\usepackage{epstopdf}
\usepackage{cite}
\usepackage{siunitx}

\usepackage{booktabs}

\usepackage{amsmath}
\usepackage{amsfonts}
\usepackage{tabularx}

\usepackage{float}
\usepackage{caption}

\usepackage{amsthm}

\newtheorem{theorem}{Theorem}

\theoremstyle{definition}
\newtheorem{definition}{Definition}

\newcommand{\lightcomment}[1]{\textcolor{gray}{#1}}

\title{\LARGE \bf

A-OctoMap: An Adaptive OctoMap for Online path planning
}

\author{ Yihui Mao$^{1}$ and Shuo Liu$^{1}$% <-this % stops a space
\thanks{$^{1}$Y. Mao and S. Liu are with the Department of Mechanical Engineering, Boston
University, Brookline, MA, USA. 
        {\tt\small \{maoyihui,liushuo\}@bu.edu}}%
}

\begin{document} 
\maketitle
\begin{abstract}
Downsampling and path planning are essential in robotics and autonomous systems, as they enhance computational efficiency and enable effective navigation in complex environments. However, current downsampling methods often fail to preserve crucial geometric information while maintaining computational efficiency, leading to challenges such as information loss during map reconstruction and the need to balance precision with computational demands. Similarly, current graph-based search algorithms for path planning struggle with fixed resolutions in complex environments, resulting in inaccurate obstacle detection and suboptimal or failed pathfinding. To address these issues, we introduce an adaptive OctoMap that utilizes a hierarchical data structure. This innovative approach preserves key geometric information during downsampling and offers a more flexible representation for pathfinding within fixed-resolution maps, all while maintaining high computational efficiency. Simulations validate our method, showing significant improvements in reducing information loss, enhancing precision, and boosting the computational efficiency of map reconstruction compared to state-of-the-art methods. For path planning, our approach enhances Jump Point Search (JPS) by increasing the success rate of pathfinding and reducing path lengths, enabling more reliable navigation in complex scenes.
\end{abstract}

\section{Introduction}
\label{sec:Introduction}

\subsection{Motivation}

Downsampling and path planning are essential in robotics and autonomous systems, improving computational efficiency and enabling reliable navigation in complex environments. Downsampling simplifies sensor data, allowing real-time processing while preserving key map features such as obstacles. Path planning ensures safe and optimal routes by avoiding collisions. Many downsampling approaches use hierarchical structures like voxel grids~\cite{rusu20113d} and OctoTrees to achieve efficiency. OctoMap~\cite{hornung2013OctoMap}, a probabilistic mapping framework built on OctoTree, incorporates occupancy information for 3-D mapping. However, such methods often struggle to capture fine geometric details. In contrast, geometry-preserving techniques offer better detail retention but suffer from lower computational efficiency~\cite{oleynikova2017voxblox,han2019fiesta}.

% In the realm in path planning, identifying safe regions and effectively navigating through complex, narrow settings is a critical research area. The ability to accurately and efficiently represent the environment spatially, while adapting map resolution to complex conditions, is crucial for enhancing path planning reliability and achieving optimal navigation outcomes.\shuo{this sentence nedds to be revised totally,there is no need to mention safe region,and no control}.

% In complex and cluttered environments filled with narrow passages, it becomes especially crucial to implement trajectory planning swiftly while maintaining effective tracking and real-time control over these trajectories (see \cite{liu2017planning, liu2023iterative, liu2024safety})\shuo{no control, revise the sentence, revise the reference}. 

In the domain of path planning, graph-based search algorithms \cite{4082128,harabor2011online} have long served as a foundational approach due to their straightforward representation of space. However, these methods are constrained by their fixed resolution, limiting their adaptability to environmental complexities. A fixed resolution often fails to capture accurate obstacle boundary details or narrow passages, making it difficult to find globally optimal paths and potentially leading to path finding failures.

To address the limitations of hierarchical structures in downsampling and fixed-resolution mapping for path planning, we utilize an optimized OctoMap, a hierarchical tree-based data structure that represents point cloud clusters and obstacles at multiple scales. This approach allows downsampling to preserve essential geometric features critical for accurate map reconstruction and provides a more flexible representation of obstacles, enhancing the optimality and feasibility of pathfinding compared to traditional fixed-resolution maps.

\subsection{Related work}

\subsubsection{Volumetric Representation of Space}

Traditional volumetric reconstruction approaches in robotics have primarily leveraged voxel-based methods, such as voxel-based techniques and hierarchical structures like OctoTrees\cite{hornung2013OctoMap}. These methods have been favored for their rapid execution, which is crucial for on-the-fly applications in dynamic environments. However, they often fall short when it comes to capturing fine surface details, as they usually sacrifice spatial resolution for computational efficiency.

 While voxel-based 3D mapping is computationally efficient, it often lacks the granularity required for high-fidelity tasks such as precise obstacle avoidance. Methods like Truncated Signed Distance Fields (TSDF) and Euclidean Signed Distance Fields (ESDF) improve distance metrics~\cite{oleynikova2017voxblox,han2019fiesta}, but still suffer from data coarseness and high memory costs, limiting their scalability~\cite{schmid2020efficient}. Neural Radiance Fields (NeRF) offer detailed 3D reconstructions~\cite{mildenhall2021nerf,rosinol2023nerf}, yet their computational demands hinder real-time use in dynamic settings. In contrast, our method adaptively refines map resolution only in geometrically complex regions, preserving fine object boundaries without uniformly increasing memory or computation, unlike traditional OctoMap or fixed-resolution voxels.

\subsubsection{Point Cloud Downsampling}

Point cloud downsampling techniques aim to balance efficiency with the preservation of geometric features. Traditional methods like Random Sampling (RS) and Farthest Point Sampling (FPS) each have distinct trade-offs. RS is simple and fast, selecting points randomly, but often fails to preserve geometric structure or ensure uniform coverage, leading to significant information loss~\cite{eldar1997farthest,moenning2003fast}. In contrast, FPS iteratively selects the farthest points from the chosen set, offering better geometric fidelity, but its high computational cost limits real-time applicability~\cite{vitter1984faster}.

Voxel-based sampling (VBS) divides space into a grid of voxels and retains a representative point per voxel, either the centroid or the nearest original point~\cite{rusu20113d,li2022psegnet}. This method offers high computational efficiency suitable for real-time applications and can be combined with RS and FPS strategies, such as Uniform Voxel Sampling (UVS) and Voxelized Farthest Point Sampling (VFPS). However, its fixed voxel size limits adaptability, leading to detail loss in high-density regions. Moreover, representing all points in a voxel with a single centroid results in constant local density and loss of geometric variation.

3-D Edge-Preserving Sampling (3DEPS) addresses these limitations by emphasizing sharp features~\cite{li2022psegnet}. Inspired by sketching techniques, it captures complex 3-D shapes more effectively by applying a 3-D Surface Boundary Filter (SBF) to separate edge points from internal points~\cite{li2022plantnet}. A new point cloud is then reconstructed by adjusting the edge-to-internal point ratio, balancing detail preservation and overall distribution. However, this ratio requires empirical tuning, increasing system complexity. We improve downsampling by constructing convex hulls from node-boundary extremal points and recursively eliminating interior points. This preserves sharp features and spatial structure more effectively than voxel or random sampling, while supporting real-time performance through parallel processing.

\subsubsection{Path Planning Algorithms}
\label{subsec: Path Planning Algorithms}
In path planning, various methods can be employed to ensure effective collision-free paths. Sampling-based algorithms, such as Rapidly-exploring Random Trees (RRT) and Probabilistic Roadmaps (PRM), achieve impressive results in high-dimensional, complex spaces by randomly expanding searches within the feasible space to find a viable path\cite{lavalle1998rapidly,kavraki1996probabilistic}. The efficiency of these methods heavily relies on the techniques used for sampling points in free regions and the approach to nearest neighbor searches.

Grid search-based algorithms, such as Dijkstra, A*, and their variants, offer resolution completeness, ensuring the shortest path can be found in grid-based searches. A*, an extension of Dijkstra, introduces heuristics to estimate path cost and improves search efficiency. Jump Point Search (JPS)~\cite{harabor2011online}, a grid-optimized variant of A*, accelerates search by pruning unnecessary nodes based on jump point logic. However, these methods are limited by fixed grid resolution, which restricts adaptability and forces a trade-off between detail and memory usage. Rather than modifying JPS itself, we enhance path planning by constructing an adaptive uniform grid aligned with OctoTree boundaries, capturing obstacle edges more precisely. This improves feasibility in narrow or irregular regions, where fixed-resolution grids may fail, leading to higher success rates and shorter paths. To address these challenges, we propose an adaptive grid mapping framework that integrates enhanced OctoMap structures with uniform grid projection and dynamic spatial refinement, enabling efficient environment representation while preserving obstacle boundaries and supporting real-time planning.

\subsection{Contributions}

We propose a novel framework incorporating a novel tree-based data structure to improve the identification of feasible solutions in congested environment. In particular, the contributions are as follows:

\begin{enumerate}
\item  We introduce an adaptive OctoMap with multiscale segmentation, dynamically adjusting resolution based on geometric and control precision requirements. This allows fine-grained representation in complex regions without global overhead.

\item  Our method enables parallel downsampling via local convex hull approximation, preserving sharp features while reducing point density and maintaining real-time feasibility.

\item Our adaptive grid mapping framework integrates with JPS, allowing it to operate on geometry-aware grids that improve both success rate and path length in complex environments.
\end{enumerate}

% \calin{there are many abbreviations throughout the paper}

\section{Preliminaries}
\label{sec:Preliminaries}
In this section, we provide background information on uniform-grid OctoMap  and Jump Point Search (JPS).

\subsection{Uniform-grid OctoMap}

\begin{figure}[h]
    \centering
    \begin{minipage}[b]{0.45\linewidth}
        \centering
        \includegraphics[width=\linewidth]{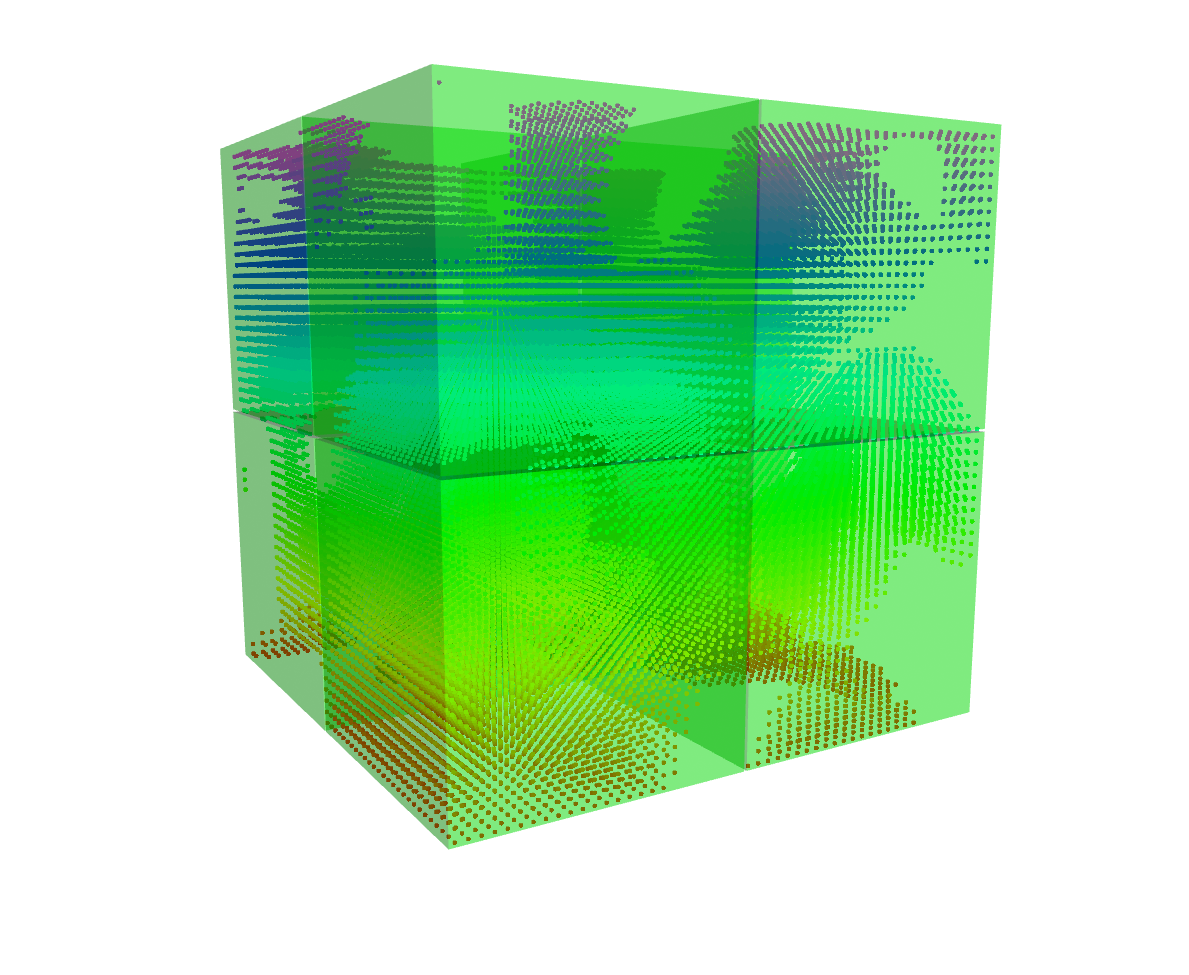}
        \caption{OctoTree map with hierarchical voxels.}
        \label{fig:image1}
    \end{minipage}
    \hspace{0.05\linewidth} % Space between the images
    \begin{minipage}[b]{0.45\linewidth}
        \centering
        \includegraphics[width=\linewidth]{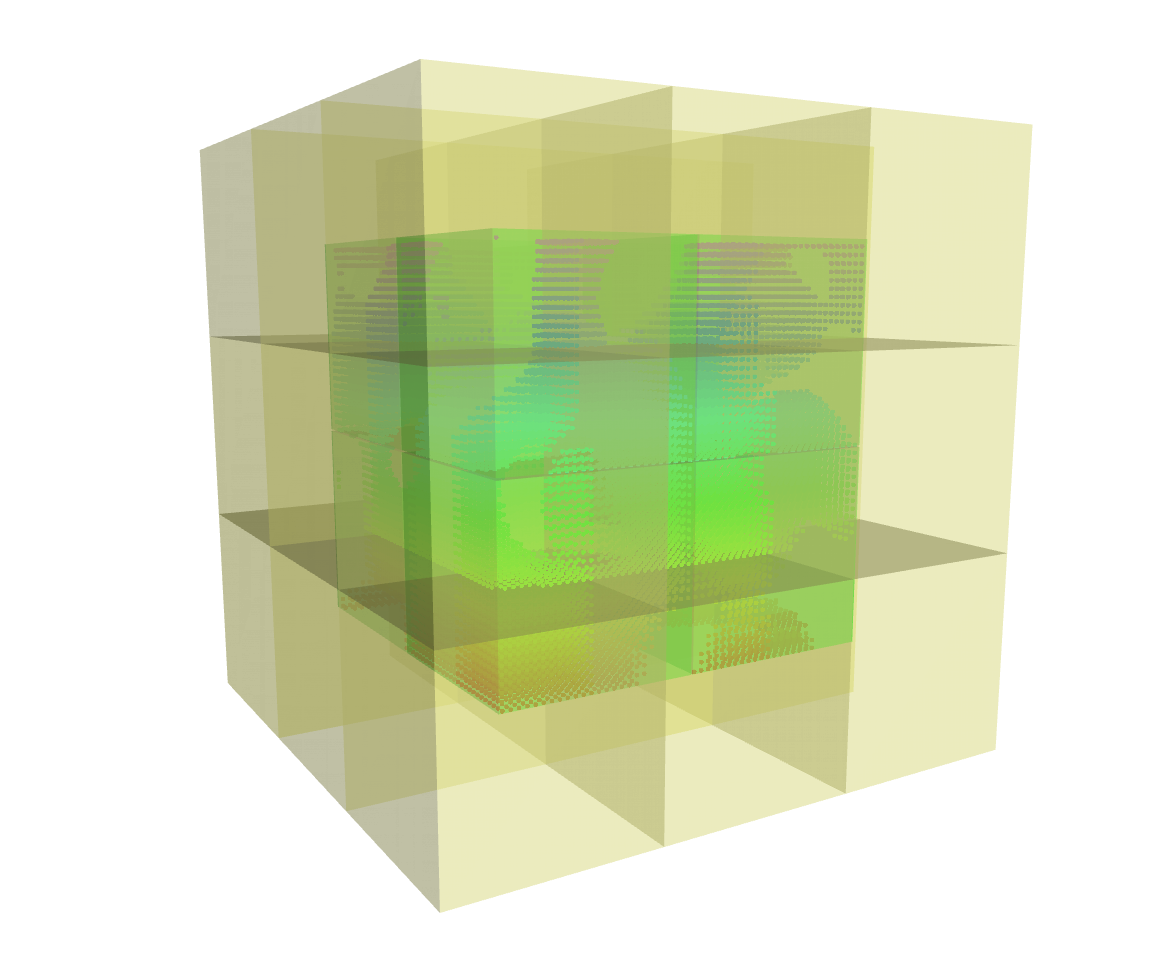}
        \caption{Uniform grid map based on Data Structure \ref{datastructure: uniform grid map}.}
        \label{fig:image2}
    \end{minipage}
\end{figure}

Although OctoMap is typically implemented as an adaptive OctoTree-based map, it can also be simplified into a uniform-resolution grid by fixing the voxel size and disabling recursive subdivision. This simplified version retains the OctoMap framework and data access methods, while allowing efficient indexing like a uniform grid. We refer to this representation as the \textbf{Uniform Grid OctoMap}, which is illustrated in Fig.~\ref{fig:image2} and formalized in \textbf{Data Structure~\ref{datastructure: uniform grid map}}.

\begin{algorithm}
\caption{Data Structure 1: Uniform Grid OctoMap}
\label{datastructure: uniform grid map}
\begin{algorithmic}[1]
\STATE \textbf{Struct} UniformGridMap
\STATE \quad \text{array} \textless int, d\textgreater grid\_map\_amount
\STATE \quad \text{multi\_array} \textless bool, d\textgreater map
\end{algorithmic}
\end{algorithm}
Figure ~\ref{fig:image1} illustrates a traditional \textit{OctoTree} structure, which recursively subdivides 3D space into hierarchical voxels. While this representation enables adaptive resolution and memory efficiency, it involves complex tree traversal during planning. In contrast, figure~\ref{fig:image2} shows a \textit{uniform grid map}, where the environment is discretized into a fixed-resolution grid. This corresponds to Data Structure \ref{datastructure: uniform grid map}, in which the map is encoded using a fixed-size integer array to define grid dimensions (\texttt{array<int, d>}) and a multi-dimensional boolean array (\texttt{multi\_array<bool, d>}) to indicate occupancy. The uniform grid structure offers faster access and simpler indexing, making it more suitable for efficient grid-based planners such as JPS.

\begin{theorem}[Pigeonhole principle~\cite{4115264}]
\label{thm:pigeonhole}
For natural numbers \(k\) and \(m\), if \(n = km + 1\) objects are distributed among \(m\) sets, the pigeonhole principle asserts that at least one of the sets will contain at least \(k + 1\) objects.
\end{theorem}

We apply Thm.~\ref{thm:pigeonhole} to the problem of uniform grid coverage. In a uniform grid map with fixed cell size \( n_s \), if a gap of width at least \( n_s \) exists in the environment, then by the pigeonhole principle, at least one grid cell boundary must intersect this gap. This ensures that any obstacle-free passage of sufficient width will be captured by the grid structure, and thus, it can be considered during planning. Therefore, the use of uniform-resolution grids is sufficient to detect narrow but traversable gaps, as long as the gap size exceeds the grid resolution.

\subsection{Jump Point Search}
A graph search algorithm can be used to find a valid path that avoids collisions with obstacle-occupied grids. Jump Point Search (JPS) is particularly suited for rapid path planning. We define a path $\pi = \langle n_i, \ldots, n_j \rangle$ as a sequence of nodes from $n_i$ to $n_j$, with its length (or cost) denoted by $\text{len}(\pi)$. Let $x$ be a node, $p(x)$ its parent, and $\text{neighbours}(x)$ the set of adjacent nodes. A specific neighbor $n$ is an element of $\text{neighbours}(x)$, and $\pi \setminus x$ denotes the path $\pi$ with node $x$ omitted.
 
In a 2-D map, each node may have up to 8 neighbors. Moving straight (horizontally or vertically) to a traversable neighbor costs 1, while diagonal moves incur a cost of $\sqrt{2}$. 

For straight moves, a neighbor $n$ is pruned if it can be reached from $x$'s parent $p(x)$ with less or equal cost without passing through $x$, i.e.,
\[
\text{len}(\langle p(x),\dots,n \rangle \setminus x) \leq \text{len}(\langle p(x),x,n \rangle).
\]
For diagonal moves, the pruning condition is stricter:
\[
\text{len}(\langle p(x),\dots,n \rangle \setminus x) < \text{len}(\langle p(x),x,n \rangle).
\]

The natural neighbors of $x$ are those remaining after pruning, assuming $\text{neighbours}(x)$ excludes obstacles. \textbf{Forced neighbors} refer to nodes that are not natural neighbors but must be explored to ensure optimality. A node $n$ is a forced neighbor if:
\begin{enumerate}
    \item $n$ is not a natural neighbor; and
    \item $\text{len}(\langle p(x),x,n \rangle) < \text{len}(\langle p(x),\dots,n \rangle \setminus x)$.
\end{enumerate}

A node qualifies as a \textbf{jump point} if it is the start or goal node, has a forced neighbor, or (in diagonal movement) leads directly to another jump point, as defined in~\cite{harabor2011online,liu2017planning}. Starting from initial node, the algorithm recursively searches for new jump points in every direction until the terminal node is found or a dead end is reached in that direction. Upon reaching the terminal node, it backtracks through the jump points to construct the shortest path, $\pi=\left \langle n_{0}, n_{1},\dots,n_{E} \right \rangle$, from start to finish. Path searching with jump point pruning has been proven to be cost-optimal in~\cite{harabor2011online}. JPS has also been shown to efficiently generate safe paths in environments with clustered dynamic obstacles, guiding the system effectively~\cite{liu2024safety}.

\section{OctoMap-Based Adaptive Grid Mapping Framework}
\label{sec:Methodology}
This section presents the proposed adaptive grid mapping framework, which builds upon the standard OctoMap representation to enable geometry-aware refinement and planning-oriented grid generation. At its core, we introduce a hierarchical data structure, referred to as the Adaptive OctoMap Tree, which supports recursive space decomposition while preserving local obstacle boundary information.
\subsection{TreeNode: Node Structure for Adaptive OctoMap Tree}
Each node in the tree represents a localized region of space and maintains geometric and topological attributes that support adaptive partitioning. In contrast to traditional OctoMap nodes, our node structure includes a local point cloud segment and two axis-aligned bounding boxes: one representing the node's physical boundary and the other capturing the position of the most recent split. The complete definition of the tree node structure is provided in \textbf{Data Structure~\ref{datastructure:octreeNode}}.

\begin{algorithm}
\caption{Data Structure 2: Node representation in the Adaptive OctoMap Tree}
\label{datastructure:octreeNode}
\begin{algorithmic}[1]
\STATE \textbf{Struct} TreeNode 
\STATE \quad \lightcomment{// Common Attributes in vanilla Octotrees} 
\STATE \quad TreeNode* Parent;
\STATE \quad TreeNode* Children[\(2^d\)];
\STATE \quad \text{vector}\textless\(\mathcal{P}\)\textgreater \text{PointCloud};
\STATE \quad \lightcomment{// New Attributes in D-octoTree}
\STATE \quad float SplitBoundry[\(2d\)]
\STATE \quad float NodeBoundry[\(2d\)]
\end{algorithmic}
\end{algorithm}
Each internal node subdivides its region into child subregions based on bisection along coordinate axes. The bisection information is encoded in two key fields:
\begin{itemize}
  \item \texttt{SplitBoundary}: the location of splitting planes along each axis.
  \item \texttt{NodeBoundary}: the spatial extent (min/max bounds) of the node region.
\end{itemize}

These attributes are used during tree construction and map conversion to precisely identify geometric boundaries of obstacles and free space.
To determine whether a node should be further subdivided, we introduce the concept of the Minimum Controllable Region (MCR). This concept establishes a principled lower bound on spatial resolution based on the agent's ability to traverse narrow regions. The MCR defines the minimal region width required for safe navigation, and serves as a stopping condition for recursive tree expansion.

\begin{definition}[Minimum Controllable Region (MCR)]
The MCR centered at a reference point \( \mathbf{p}_0 \in \mathbb{R}^d \) is defined as
\begin{equation}
\label{MCR}
\text{MCR} = \left\{ \mathbf{p} \in \mathbb{R}^d \mid \|\mathbf{p} - \mathbf{p}_{0}\|_\infty \leq  \epsilon_{\max} \right\},
\end{equation}
where \( \epsilon_{\max} \) denotes the maximum allowable deviation in any dimension.
\end{definition}
Here, \( \|\cdot\|_\infty \) denotes the infinity norm, i.e., the maximum absolute difference across dimensions. The value \( \epsilon_{\max} \) is typically determined by the system's localization uncertainty, physical footprint, or clearance requirement. This region characterizes the smallest space where an agent can reliably operate without further subdivision. To determine the required tree depth for a given axis-aligned map size \( L \), we use:
\begin{equation}
\label{eq:depth}
\text{depth} = \left\lceil \log_2 \left( \frac{L}{k \cdot p} \right) \right\rceil,
\end{equation}
where \( p \) is the MCR size along that axis, and \( k > 1 \) is a safety scaling factor (e.g., \( k=2 \)) to ensure sufficient resolution.

\subsection{Adaptive OctoMap Tree Structure}
Building on the \texttt{TreeNode} definition, we now formalize the structure of the entire adaptive tree used for recursive spatial subdivision. To enable traversal, refinement, and global reasoning, the tree is represented both structurally as a hierarchical object and formally as a graph.

Formally, we define the tree as a directed acyclic graph (DAG) where each node stores geometric, hierarchical, and relational information. This abstraction supports efficient representation and manipulation of map regions during adaptive mapping and planning.

\begin{definition}
An adaptive OctoMap Tree $T$ is defined as a graph $T = (V, E)$ where:
\begin{itemize}
  \item $V = \{v_1, v_2, \ldots, v_n\}$ is the set of nodes,
  \item $E = \{(v_i, v_j) \mid v_j \text{ is a child of } v_i\}$ represents parent-child edges.
\end{itemize}
Each node $v_i$ is defined as a tuple $(\text{value}, \text{parent}, \text{children})$, where \texttt{value} stores its point cloud identifier, and \texttt{parent}/\texttt{children} encode hierarchical connectivity.
\end{definition}

To support efficient access and recursive processing, we define the tree using the structure shown in \textbf{Data Structure~\ref{datastructure:octree}}. It includes a root node and a tree depth, as determined by the MCR criterion in Eqn.~\eqref{eq:depth}.
\begin{algorithm}
\caption{Data Structure 3: OctoTree}
\label{datastructure:octree}
\begin{algorithmic}[1]
\STATE \textbf{Struct} OctoTree 
\STATE \quad TreeNode* root;
\STATE \quad int Depth;
\end{algorithmic}
\end{algorithm}

\subsection{Tree Construction}
\setcounter{algorithm}{0}
With the structure of the adaptive OctoMap Tree formally defined, we now describe the process of constructing this tree from a raw point cloud input. The goal is to recursively subdivide space based on geometric content and the MCR criterion until a resolution limit is reached.
The tree construction process is recursive. Starting from a root node, the point cloud is progressively divided into smaller subregions. For each node, we first check if it satisfies the termination condition based on depth or obstacle sparsity. If not, we bisect the region along its longest axis, update the \texttt{SplitBoundary} and \texttt{NodeBoundary}, allocate points to child nodes, and repeat the process. The recursion ends when the region size reaches the MCR threshold or the depth limit is met. The full procedure is detailed in \textbf{Algorithm~\ref{algo:octree_build}}.

\begin{algorithm}
\caption{Algorithm 1: OctoTree Construction}
\label{algo:octree_build}
\begin{algorithmic}[1]
\STATE \textbf{function} T. Build($\mathcal{P}$) 
\STATE \textbf{Input}: $\mathcal{P}$, $\mathcal{L}$
\FOR{$p \in \mathcal{P}$}
    \STATE T. push\_point\_into\_tree($p$, $\mathcal{L}$)
\ENDFOR
\end{algorithmic}

  \hrulefill % Horizontal line      

% \caption{Push Point into Tree}
\begin{algorithmic}[1]
\STATE \textbf{function} T. push\_point\_into\_tree(curNode, $p$, $\mathcal{L}$)
    \STATE \textbf{Input}: $p$ \texttt{// Point}
    
    \STATE  OctoTreeNode curNode $\leftarrow$ root
    \STATE curNode. depth $\leftarrow$ maxDepth
     \WHILE{curNode. depth $>$ 0}
        \STATE  bisects\_each\_level\_along\_axis()
        \STATE idx $\xleftarrow{}$allocate and store points to child nodes
        \IF{curNode. depth $==$ 0}
            \STATE $\mathcal{L}$. push\_back(curNode)
            \STATE \textbf{return}
        \ELSE
            \STATE construct\_node(curNode, idx)
            \STATE curNode $\leftarrow$ curNode. children[idx]
            \STATE push\_point\_into\_tree(curNode, $p$, $\mathcal{L}$)
        \ENDIF
    \ENDWHILE
\label{alg: octree_build}
\end{algorithmic}
\end{algorithm}

\subsection{Uniform Grid Mapping}

\begin{figure}[H]
    \centering
    \includegraphics[width=\linewidth]{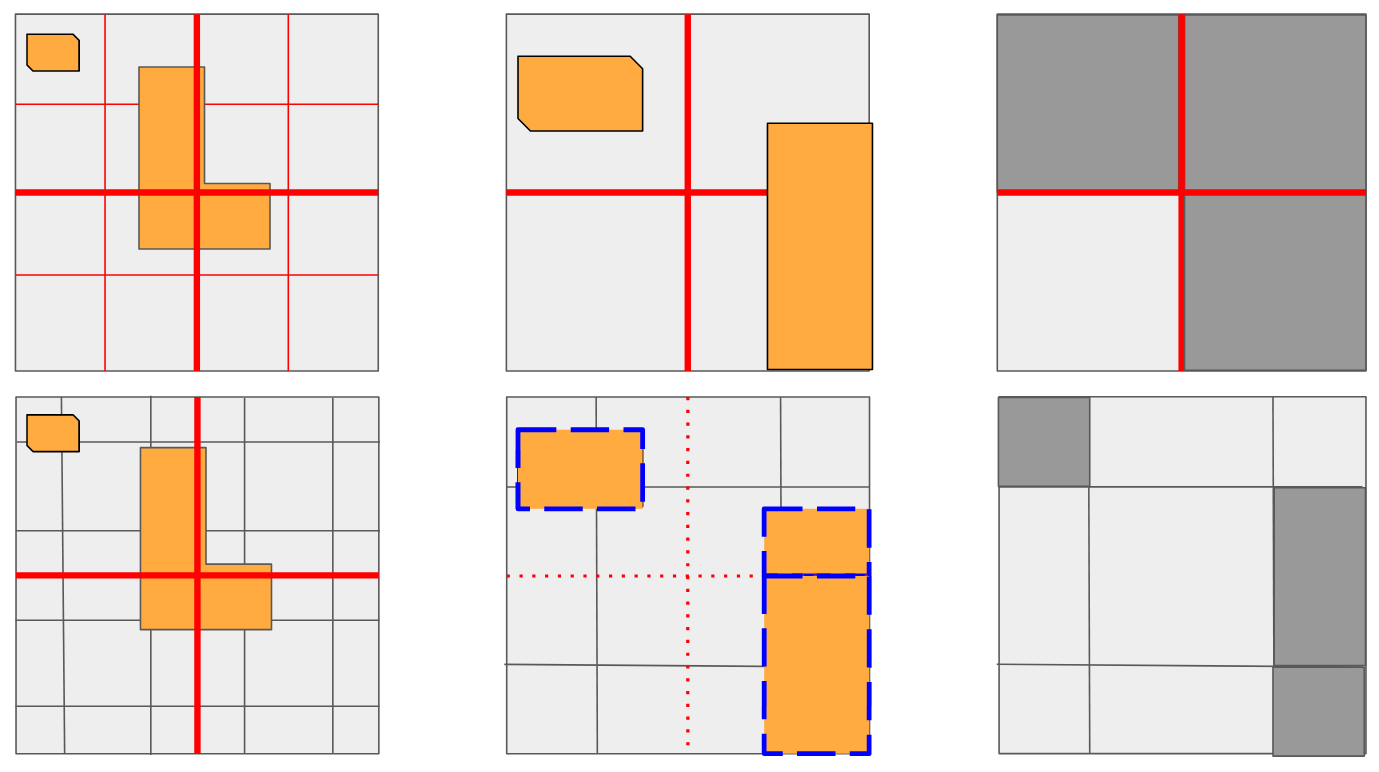}
    \caption{Illustration of mapping from OctoTree to uniform grid. The top row shows the fixed uniform grid method, while the bottom row shows our adaptive method. Red lines indicate OctoTree splits at different levels; red dotted lines represent \texttt{SplitBoundary}, and blue dotted lines represent \texttt{NodeBoundary}. The right columns visualize the resulting grid structures under both methods.}
    \label{fig:octree_mapping}
\end{figure}

As shown in Fig.~\ref{fig:octree_mapping}, we propose a novel mapping method that does not rely solely on fixed-size grids. Instead, grid occupancy is determined by the obstacle's spatial boundaries, with each grid cell aligned to the \texttt{SplitBoundary} extracted from the tree structure. The edge length of each grid equals that of the MCR, given by $2^{-\text{depth}} \cdot L$, where $L$ is the length of the map domain.

According to Thm.~\ref{thm:pigeonhole}, the MCR-based partition ensures that each cell intersects the gap between obstacles, enabling adaptive refinement near object boundaries. Unlike traditional uniform grids that may misalign with obstacle edges, our method guarantees geometric consistency between the grid layout and the map geometry, thus improving representation fidelity without increasing the total number of cells.

A cell is marked as occupied based on the relative distance between its \texttt{SplitBoundary} and \texttt{NodeBoundary}. Since the grid size is derived from the tree depth, it inherits the resolution characteristics of the original node, maintaining efficiency in memory and computation.

\subsection{Dynamic Partition}
We use JPS as the default planner operating on the adaptive grid. During planning, the system interacts with the tree structure to query occupancy; if no feasible path is found, dynamic refinement is triggered. Building on the adaptive mapping framework, we incorporate a dynamic partitioning strategy to handle cases where static uniform grids fail—particularly in environments with emerging obstacles or narrow passages.

When the planner is unable to find a feasible path, the relevant leaf nodes are identified and recursively subdivided to locally increase resolution. New point cloud data can be inserted into the refined nodes, allowing obstacle information to remain up to date. This process continues until a feasible solution is found or the resolution limit is reached.

The dynamic refinement allows the map to adjust to environmental complexity in a localized manner, avoiding the need for full map reconstruction. The recursive update logic is implemented in \textbf{Algorithm~\ref{algo:Dynamic partition}}.

\begin{algorithm}
\caption{Algorithm 2: Dynamic partition}
\label{algo:Dynamic partition}
\begin{algorithmic}[1]
\STATE \textbf{function}  dynamic\_partition( $\mathcal{L}$)
    \STATE \textbf{Input}: $\mathcal{L}$\texttt{//vector of leaf nodes' address}
    \STATE update tree depth
    \FORALL {$\mathcal{L}$}
         \STATE l $\xleftarrow{} \mathcal{L}$. pop\_front()
        \IF{ l is not \textbf{nullptr}}
        \FOR{$p \in \mathcal{P}$}
            \STATE push\_point\_into\_tree(this, $p$, $\mathcal{L}$) 
        \ENDFOR
        \ENDIF
    \ENDFOR

\end{algorithmic}
\end{algorithm}

\subsection{Downsampling}
To reduce point cloud density while preserving geometric features, we propose a structure-aware downsampling method based on local convex hull approximation. Each tree node maintains a local point cloud and operates independently, enabling parallel processing during tree construction.

For each leaf node, we first compute extremal points along each axis to form a convex polyhedron that approximates the surface geometry. Internal points are eliminated, while only representative boundary points are retained to preserve key features. This process is outlined in \textbf{Algorithm~\ref{algo: Downsampling}}.

% \begin{figure}[H]
%     \centering
%     \begin{subfigure}[b]{0.45\linewidth}
%         \centering
%         \includegraphics[width=\linewidth]{figures/new_original___.png}
%         \caption{Original method}
%         \label{fig:common_method}
%     \end{subfigure}
%     \hfill
%     \begin{subfigure}[b]{0.45\linewidth}
%         \centering
%         \includegraphics[width=\linewidth]{figures/new_downsampling___.png}
%         \caption{Our method}
%         \label{fig:our_method}
%     \end{subfigure}
%     \caption{Illustration of downsampling in a real-world map}
%     \label{algo:octree_mapping}
% \end{figure}

\begin{algorithm}
\caption{Algorithm 3: Downsampling}
\label{algo: Downsampling}
\begin{algorithmic}[1]

\STATE \textbf{function} convexifyPointCloud()
\STATE \textbf{Input}: branch \hfill \texttt{//  branch node}
\STATE \hfill convex\_vectice \hfill \texttt{//  extreme points}
\STATE \quad externalPts $\leftarrow$ branch. nodeBoundry
\STATE \quad hex $\leftarrow$ convex\_hull(externalPts)

% \STATE \quad \textcolor{gray}{// remove redundant points and assign them}
% \STATE \quad \textcolor{gray}{  // to corresponding extreme faces}
\STATE \quad Pts $\leftarrow \{ x \in \text{branch. PointCloud} \mid x \notin \text{HexRegion}\}$
\STATE \quad pts\_array $\leftarrow$ range\_search(pts)
\FOR{$k \in \{0, ..., 2^d\}$}
    \STATE \quad extreme\_pts $\leftarrow$ convex\_hull(pts\_array[k])
    \STATE \quad convex\_vertice. push\_back(extreme\_pts)
\ENDFOR
\STATE \quad \textbf{return} convex\_vertice
\end{algorithmic}
\end{algorithm}
After filtering, the retained points are used to reconstruct the mesh using the QuickHull algorithm \cite{barber1996quickhull} and triangulated via Computational Geometry Algorithms Library (CGAL) \cite{fabri2009cgal}. The resulting surface meshes are stored in the respective tree nodes and can be used for downstream planning or visualization.
\subsection{Complexity Analysis}
To better understand the computational efficiency of each module in our proposed framework, we provide a complexity analysis covering the main components: tree construction, uniform (grid) mapping, dynamic partitioning, and downsampling. 

Table~\ref{tab:complexities} summarizes the time complexities in both best-case and worst-case scenarios, with respect to the number of points, tree depth, and dimensionality of the map.
\begin{table}[H]
\centering
\begin{tabular}{|c|c|c|c|}
\hline
\textbf{Algorithm} & \textbf{Big O Time} & \textbf{Best Case} & \textbf{Worst Case } \\
\hline
Tree Construction & $O(n \log m)$ & $ O(n \log m)$  & $ O(n \log m)$ \\
Uniform Mapping & $O(k m)$ & $O(k)$ & $O(k m)$ \\
Dynamic Partition &$O\left( 2^d m\right)$ & $O\left( 2^d\right)$ & $O(2^d \cdot m)$   \\
Downsampling & $O\left(n \log \frac{\hat{n}}{m}\right)$ &  $O\left(\hat{n} \log \frac{\hat{n}}{m}\right)$ & $O( \frac{\hat{n}^2}{m})$ \\
\hline
\end{tabular}
\caption{Time complexities of different algorithms.}
\label{tab:complexities}
\end{table}

\subsection*{Notations:}
\begin{itemize}
    \item \( d \): The dimension of the map.
     \item \( k \): The constant value $k$ in the Eqn. \eqref{eq:depth}.
    \item \( n \): Total number of points to be mapped.
     \item \( \hat{n} \): Total number of points after elimination.
    \item \( m \): Total number of leaf nodes.
\end{itemize}

In Tree Construction, the time complexity is the same as the original OctoTree build in average and worst cases. Our tree covers the entire map and maintains balance, preventing points from being inserted into a single path to one leaf node. Both the Uniform (Grid) Mapping and Dynamic Partition are linear algorithms, dependent on the number of leaf nodes.

For Downsampling, we store the point cloud in each corresponding leaf node. The best and average cases of the Quickhull method are linearithmic time, while the worst case is quadratic. In our method, points inside the external face are eliminated, resulting in a computation time of \(O\left(\frac{\hat{n}}{m} \log \frac{\hat{n}}{m}\right)\) for each leaf node. The total time complexity is \(O\left(\hat{n} \log \frac{\hat{n}}{m}\right)\) in the best and average cases, and \(O\left(\frac{\hat{n}^2}{m}\right)\) in the worst case. These results highlight that each module of the framework is designed to maintain efficiency even under worst-case conditions, with tree construction and downsampling being the most computationally intensive components due to their recursive and geometry-aware nature.

\section{Case Studies}
\begin{figure*}[t]
\centering
% 第一行图片
\begin{subfigure}{.24\textwidth}
  \centering
  \includegraphics[width=\linewidth]{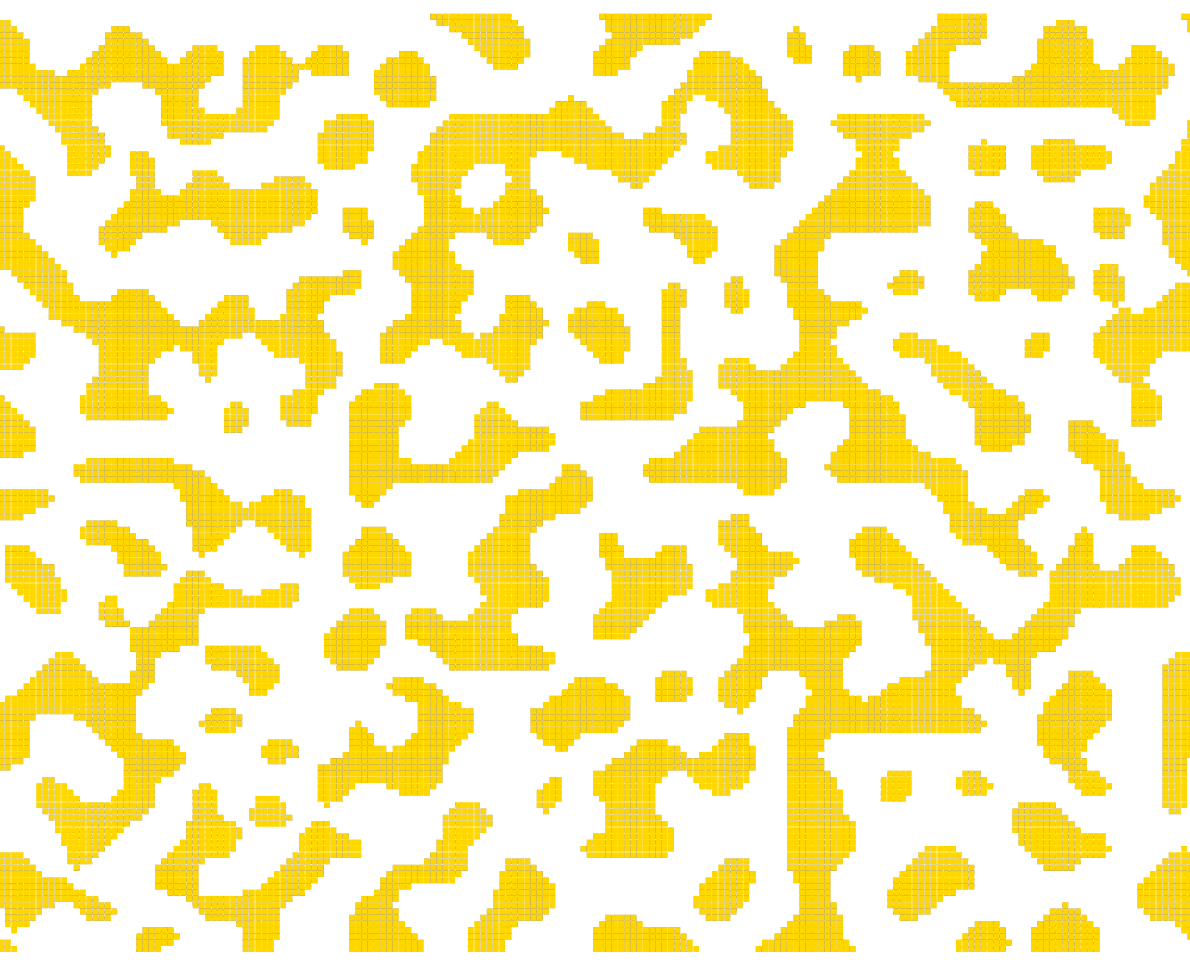}
 \caption{A 2-D point cloud map built using Perlin Noise }
\end{subfigure}
\begin{subfigure}{.24\textwidth}
  \centering
  \includegraphics[width=\linewidth]{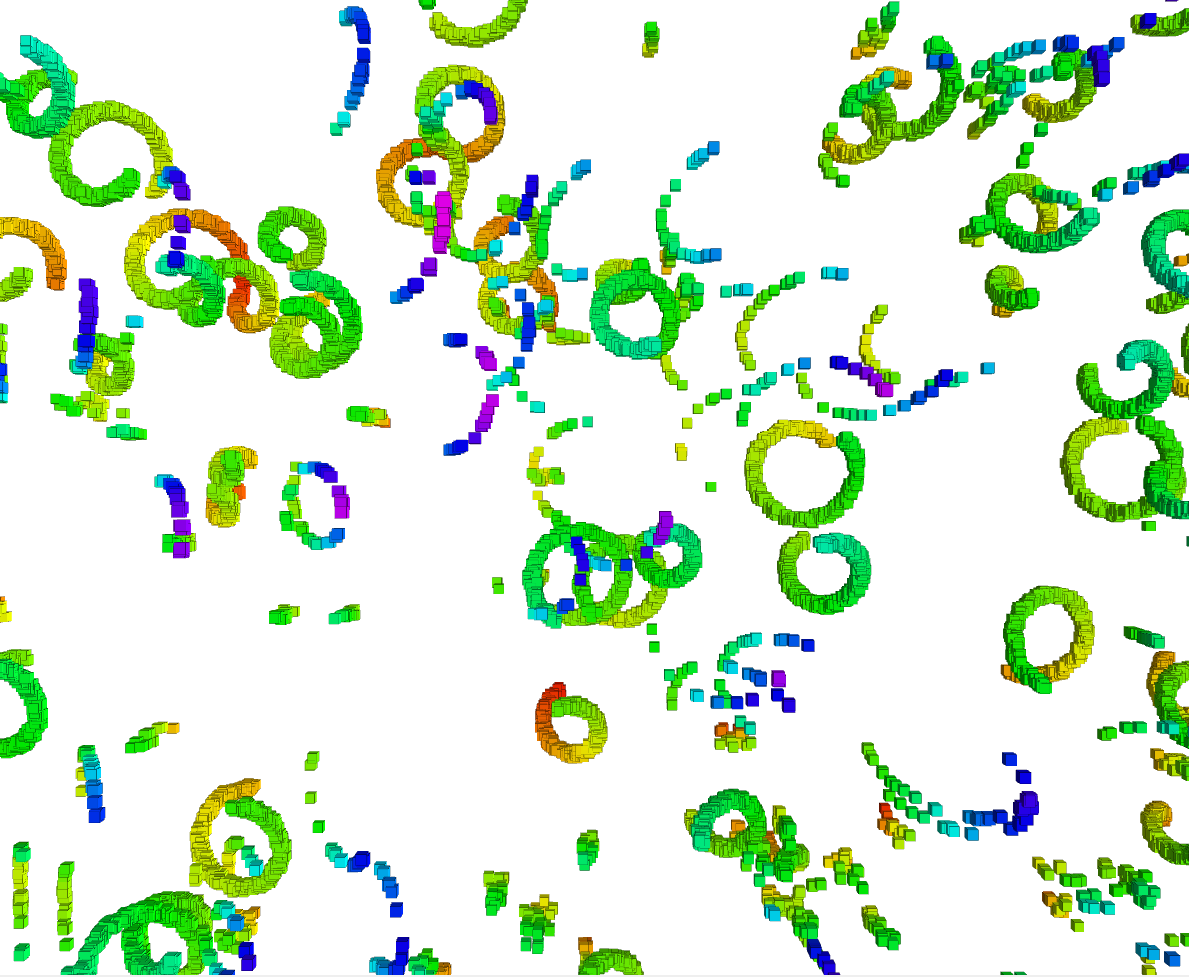}
  \caption{A 3-D point cloud map composed of non-convex geometric shapes }
\end{subfigure}
\begin{subfigure}{.24\textwidth}
  \centering
  \includegraphics[width=\linewidth]{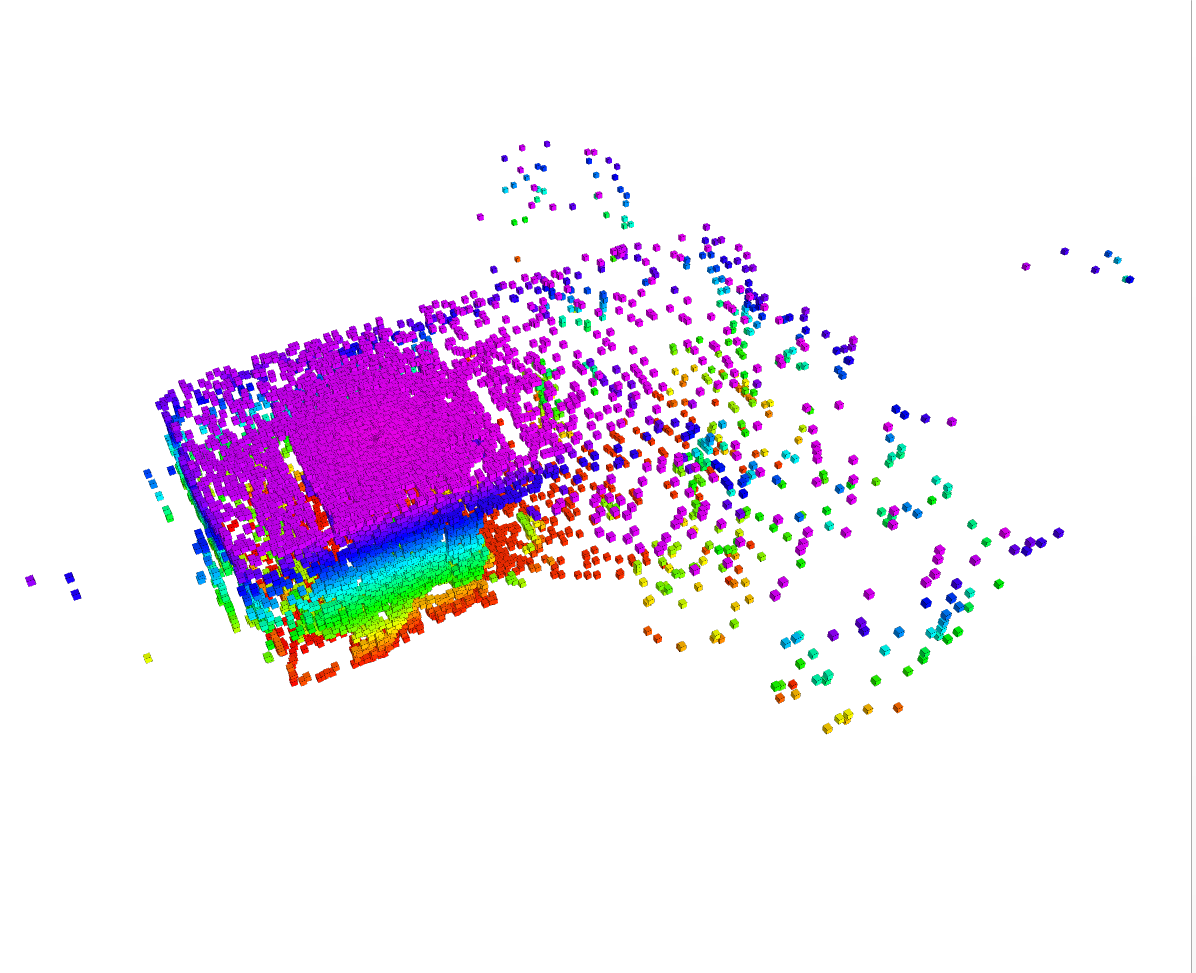}
  \caption{A 3-D point cloud map of a real-world room scan (Room 1)}
\end{subfigure}
\begin{subfigure}{.24\textwidth}
  \centering
  \includegraphics[width=\linewidth]{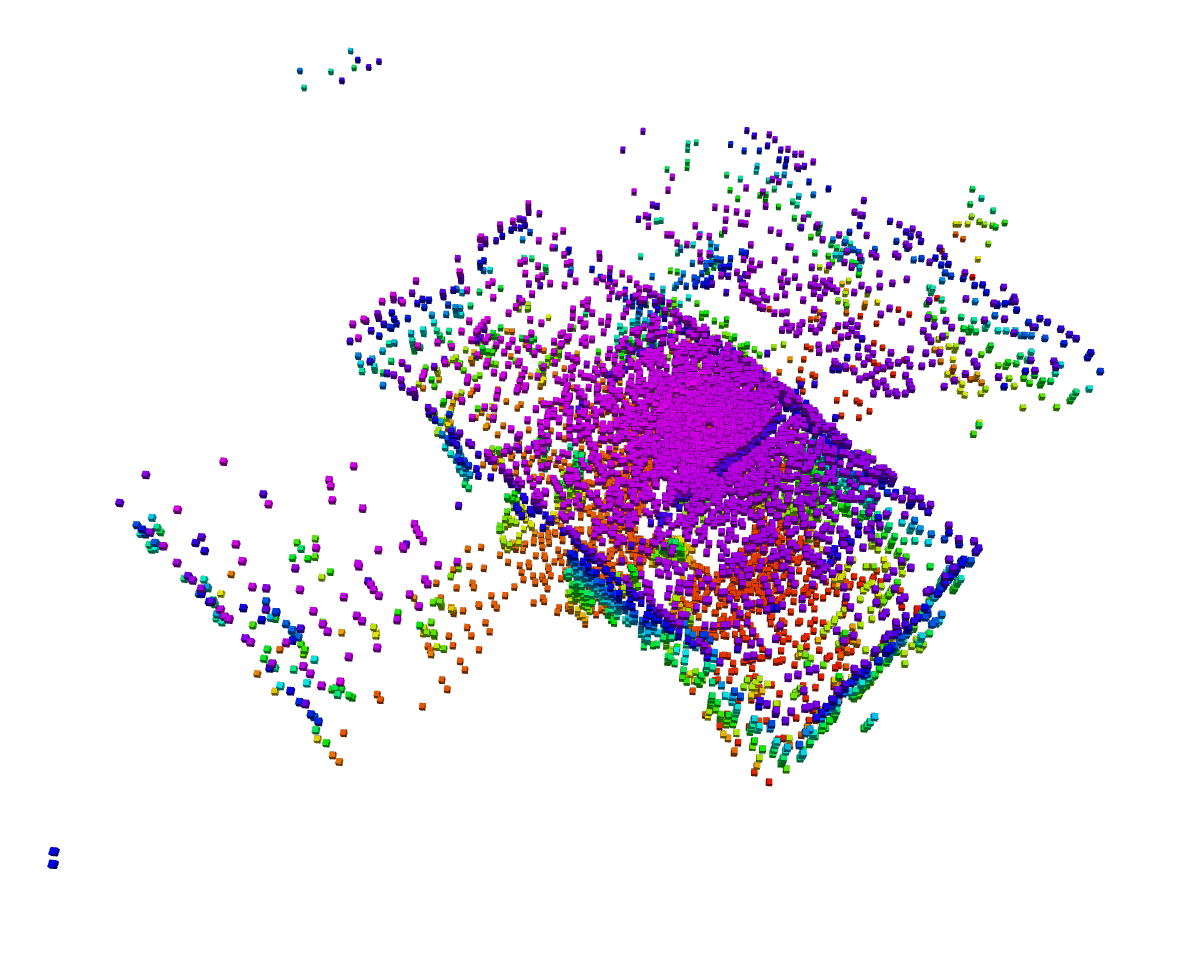}
  \caption{A 3-D point cloud map of a real-world room scan (Room 2)}
\end{subfigure}
\caption{Four experimental scenarios used to evaluate the effectiveness of the proposed framework.}
\label{fig:4_scenarios}
\end{figure*}

To evaluate the proposed framework under various environmental conditions, we design four representative scenarios involving both synthetic and real-world point cloud maps. These scenarios include 2D Perlin-noise environments, cluttered non-convex geometries, and 3D reconstructions of real indoor spaces. Figure ~\ref{fig:4_scenarios} presents the input maps used throughout the following case studies, which are based on the RViz in Robot Operation System (ROS) Noetic Ninjemys.  We used a Linux desktop with Intel Core i9-13900H running c++ for all computations. Each of the subsequent subsections focuses on a specific module of our framework, analyzed in the context of these representative environments.

 The first scenario is a 2-D map with 35 thousand of point cloud built using Perlin noise, featuring a large number of narrow passages. This setup is designed to test the performance of our algorithm in cluttered environments. The second scenario involves an artificially generated scene with non-convex geometric shapes made from point clouds, including arches, helices, cylinders, and cuboids. The third and fourth scenario are based on real-world point cloud maps of rooms, 
referred to as Room 1 and Room 2,  which respectively measures 30$\times$15$\times$4 meters and 26$\times$22$\times$4 meters, and both contains approximately 110,000 points.  These scenarios are used to validate the advanced capabilities of our algorithm through corresponding experiments.

\subsubsection{Tree Construction}
The table \ref{table:tree_build} shows the performance of the tree construction in the four scenarios.
\begin{table}[H]
\centering
\begin{tabular}{@{}ccccc@{}}
\toprule
\textbf{Dim} & \textbf{2-D} &  \textbf{3-D} &  \textbf{3-D} &  \textbf{3-D} \\
\midrule
\textbf{Scene} & \textbf{Perlin Noise}  & \textbf{Geometric Shapes} & \textbf{Room1} & \textbf{Room2} \\ \midrule
\textbf{Depth} & \textbf{Time}      & \textbf{Time}        & \textbf{Time}  & \textbf{Time}  \\ \midrule
4    & 2.47 ms          & 2.45 ms              & 4.56 ms        & 6.68 ms        \\ 
5    &2.56 ms         & 3.49 ms              & 5.91 ms        & 7.57 ms        \\ 
6    & 3.71 ms          & 4.72 ms              & 7.73 ms        & 9.01 ms        \\ 
7     &5.04 ms         & 10.86 ms             & 11.08 ms       & 14.41 ms       \\ 
8     &6.73 ms         & 19.07 ms             & 26.25 ms       & 21.17 ms       \\ \bottomrule
\end{tabular}
\caption{Time measurements for different levels and scenes.}
\label{table:tree_build}
\end{table}

\subsubsection{Uniform (Grid) Mapping and Dynamic Partition}
The Table \ref{tab:mapping} presents the performance of uniform (grid) mapping and dynamic partition at specific depths of the OctoMap. In both uniform mapping and dynamic partition, the execution times are all in milliseconds, ensuring their viability for online applications.

\begin{table}[H]
\centering
\small % Adjust font size to be smaller
\begin{tabular}{@{}ccc@{}}
\toprule
\textbf{  Algorithm  } & \textbf{  Uniform Mapping  } & \textbf{  Dynamic Partition } \\ \midrule
\textbf{Depth}     & \textbf{Time}            & \textbf{Time}              \\ \midrule
4                  & 0.009 ms                  & 1.551 ms                    \\ 
5                  & 0.021 ms                  & 1.566 ms                    \\ 
6                  & 0.044 ms                  & 1.582 ms                    \\ 
7                  & 0.069 ms                  & 1.621ms                   \\ 
8                  & 0.128 ms                  & 1.656 ms                   \\ \bottomrule
\end{tabular}
\caption{Run time of uniform mapping and dynamic partition.}
\label{tab:mapping}
\end{table}

\subsubsection{Downsampling and Mesh Triangulation} 
\begin{figure}[H]
\centering
% 第一行图片
\begin{subfigure}{.21\textwidth}
  \centering
  \includegraphics[width=\linewidth]{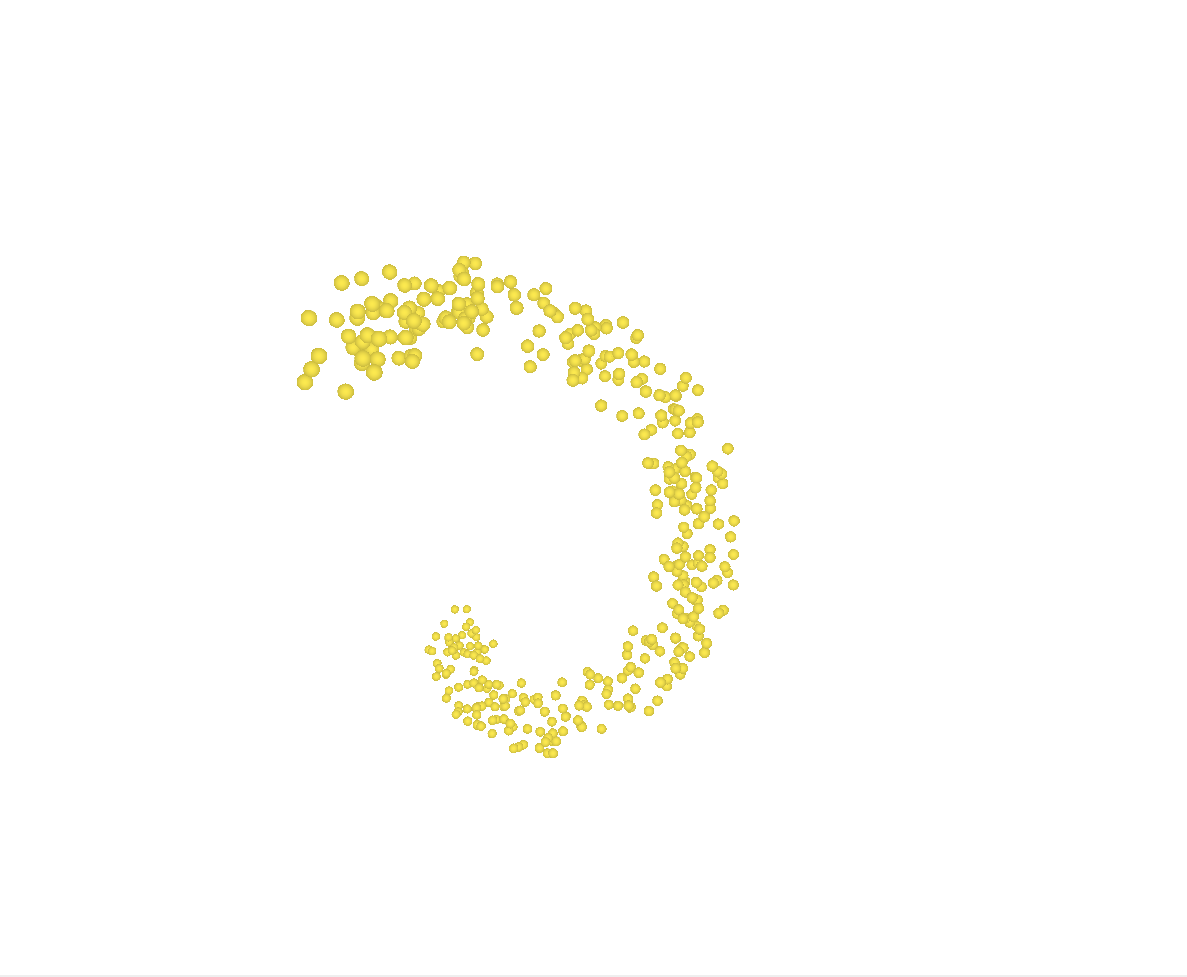}
  % \caption{A}
\end{subfigure}
\begin{subfigure}{.21\textwidth}
  \centering
  \includegraphics[width=\linewidth]{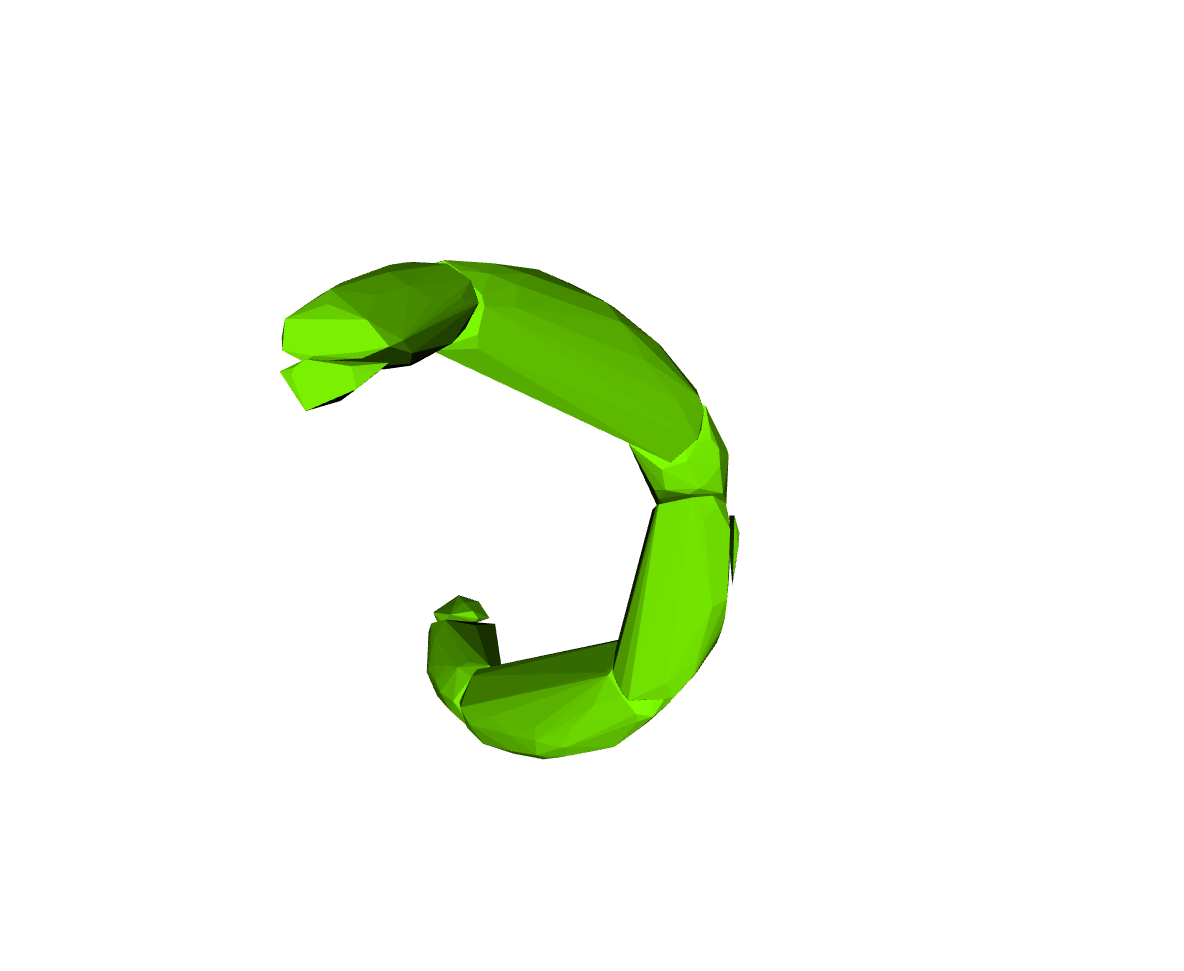}
  % \caption{B}
\end{subfigure}
\begin{subfigure}{.21\textwidth}
  \centering
  \includegraphics[width=\linewidth]{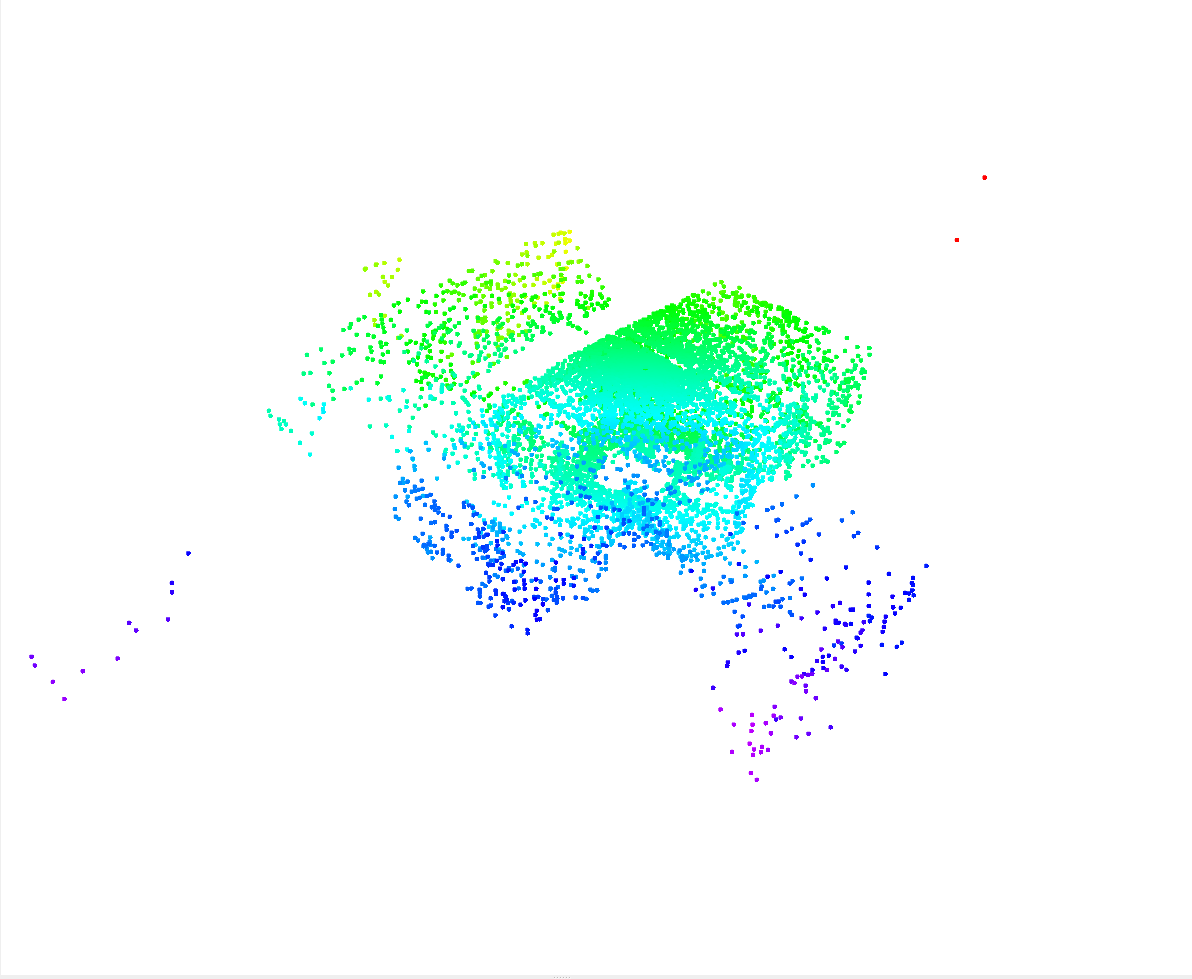}
  % \caption{A}
\end{subfigure}
\begin{subfigure}{.21\textwidth}
  \centering
  \includegraphics[width=\linewidth]{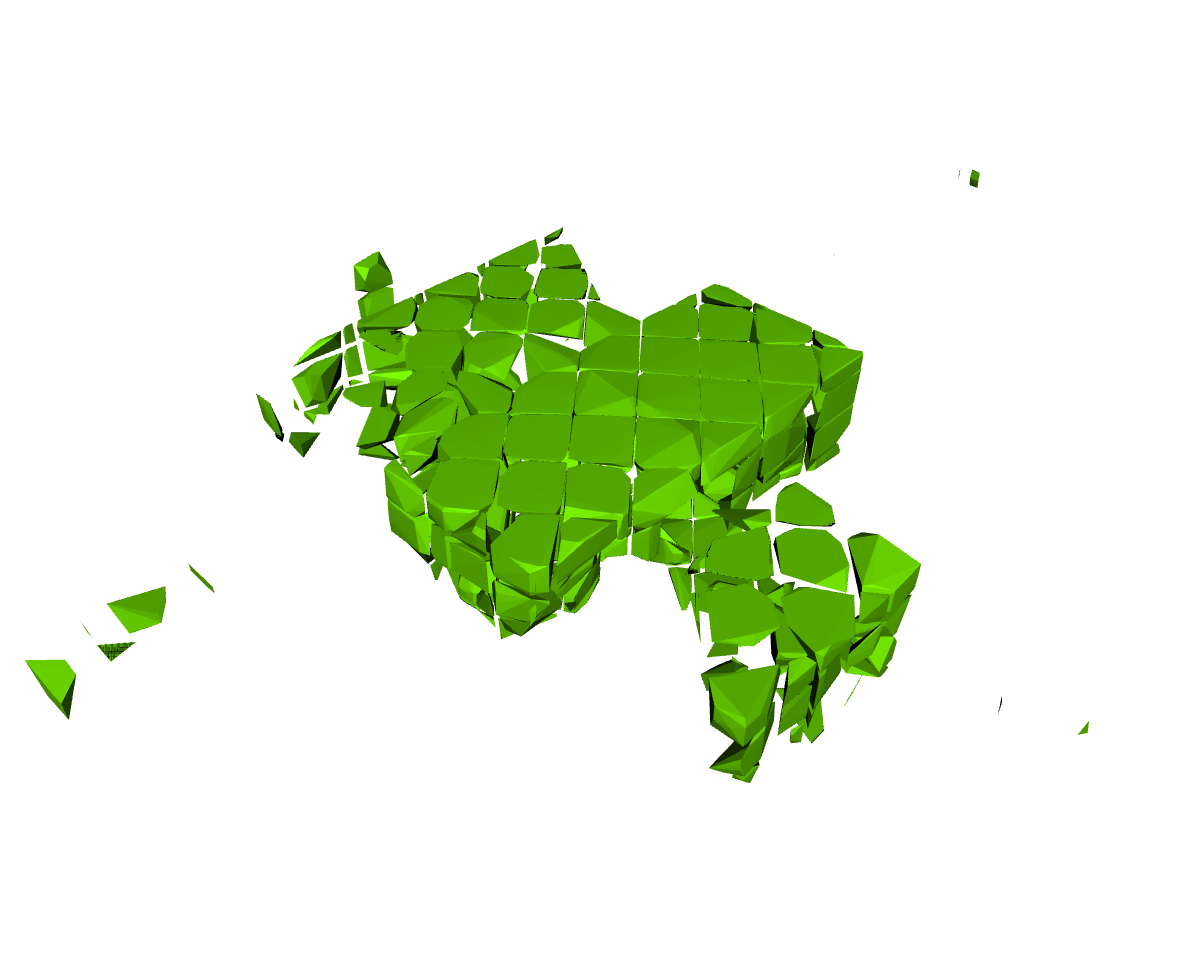}
  % \caption{B}
\end{subfigure}
\caption{Illustration of mesh triangulation after downsampling. }
\label{fig:mesh_triangulation}
\end{figure}

\begin{figure*}[t]
    \centering
    \begin{subfigure}[b]{0.32\linewidth}
        \centering
        \includegraphics[width=\linewidth]{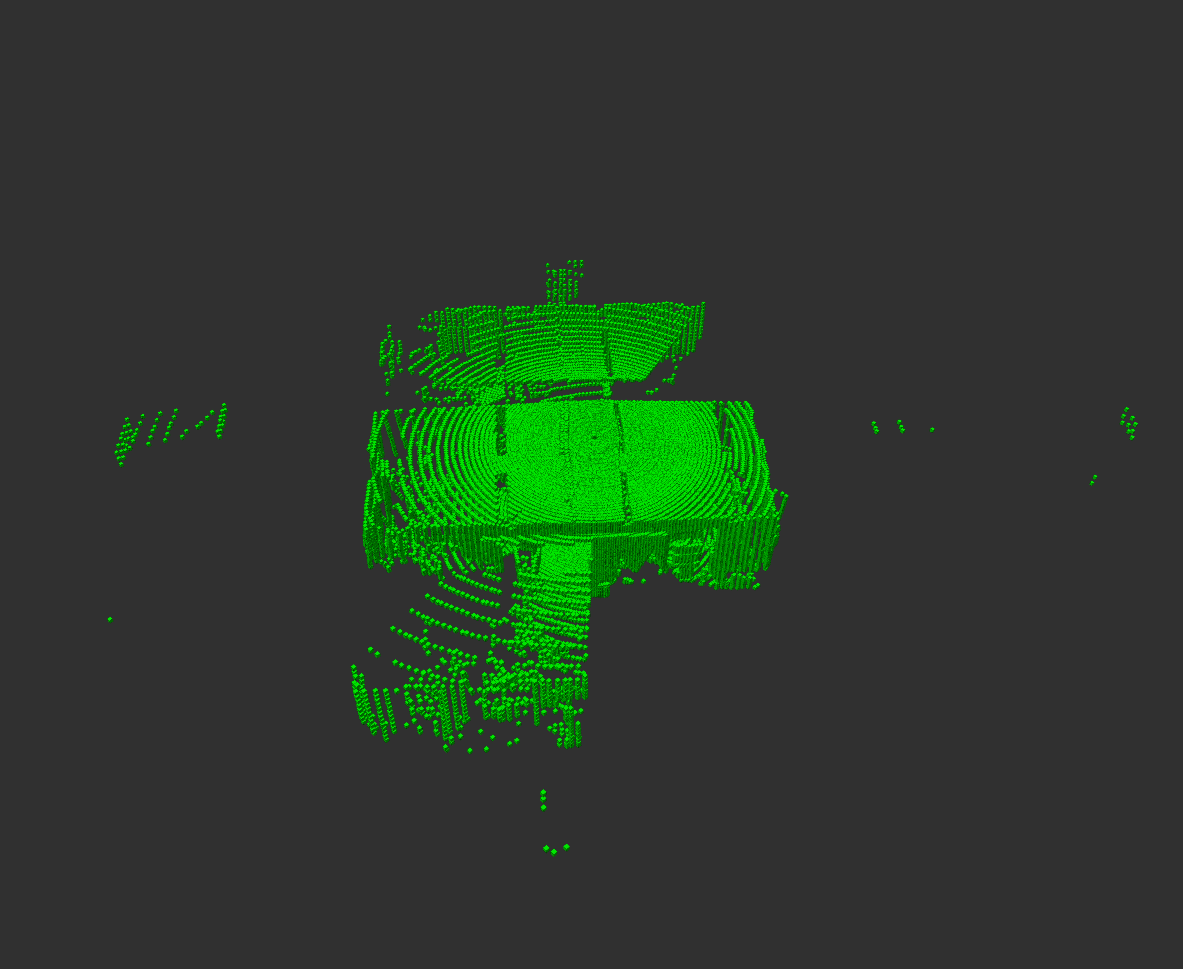}
        \caption{Original map}
        \label{fig:basic_method}
    \end{subfigure}
    \hfill
    \begin{subfigure}[b]{0.32\linewidth}
        \centering
        \includegraphics[width=\linewidth]{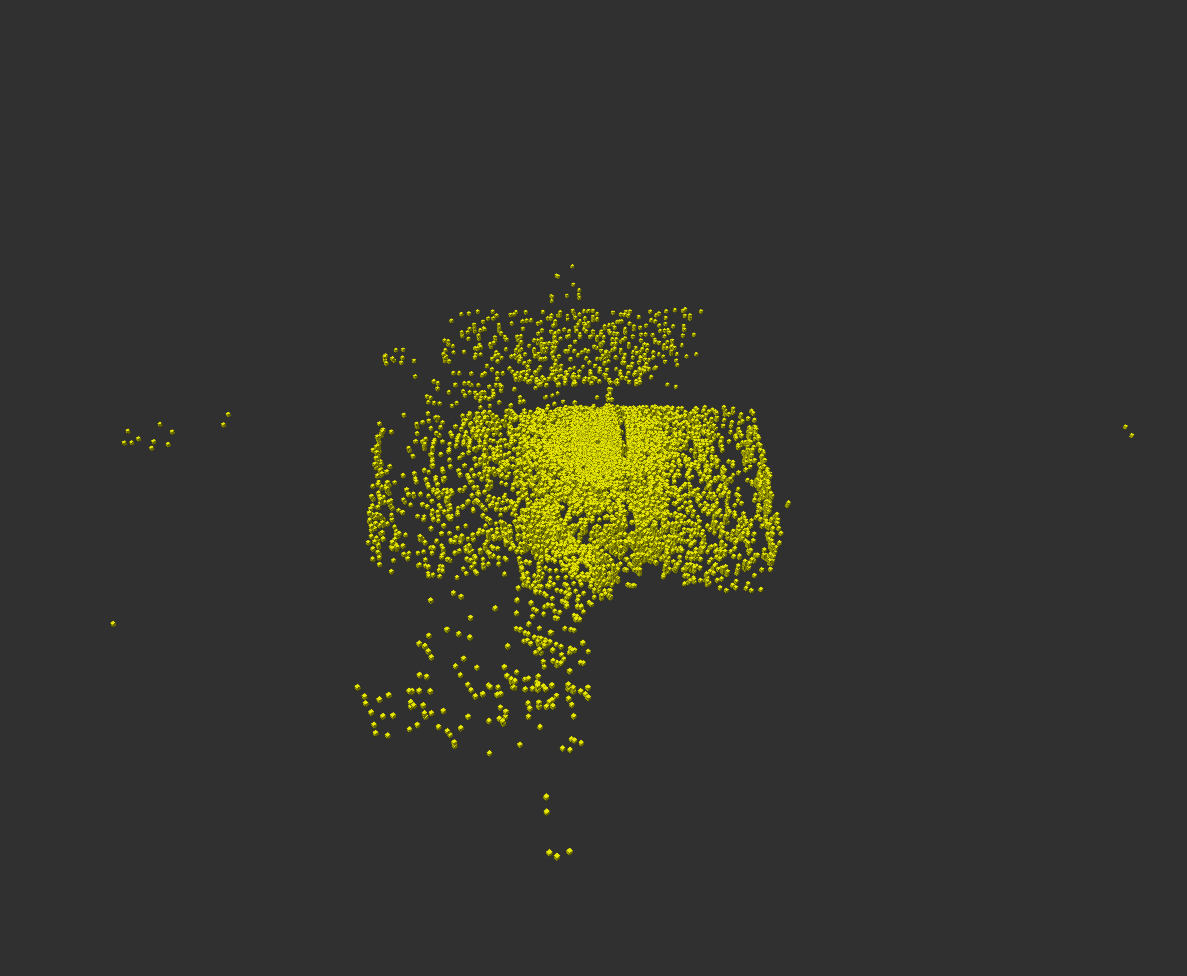}
        \caption{VoxelGrid method}
        \label{fig:Voxel_method}
    \end{subfigure}
 \hfill
        \begin{subfigure}[b]{0.32\linewidth}
        \centering
        \includegraphics[width=\linewidth]{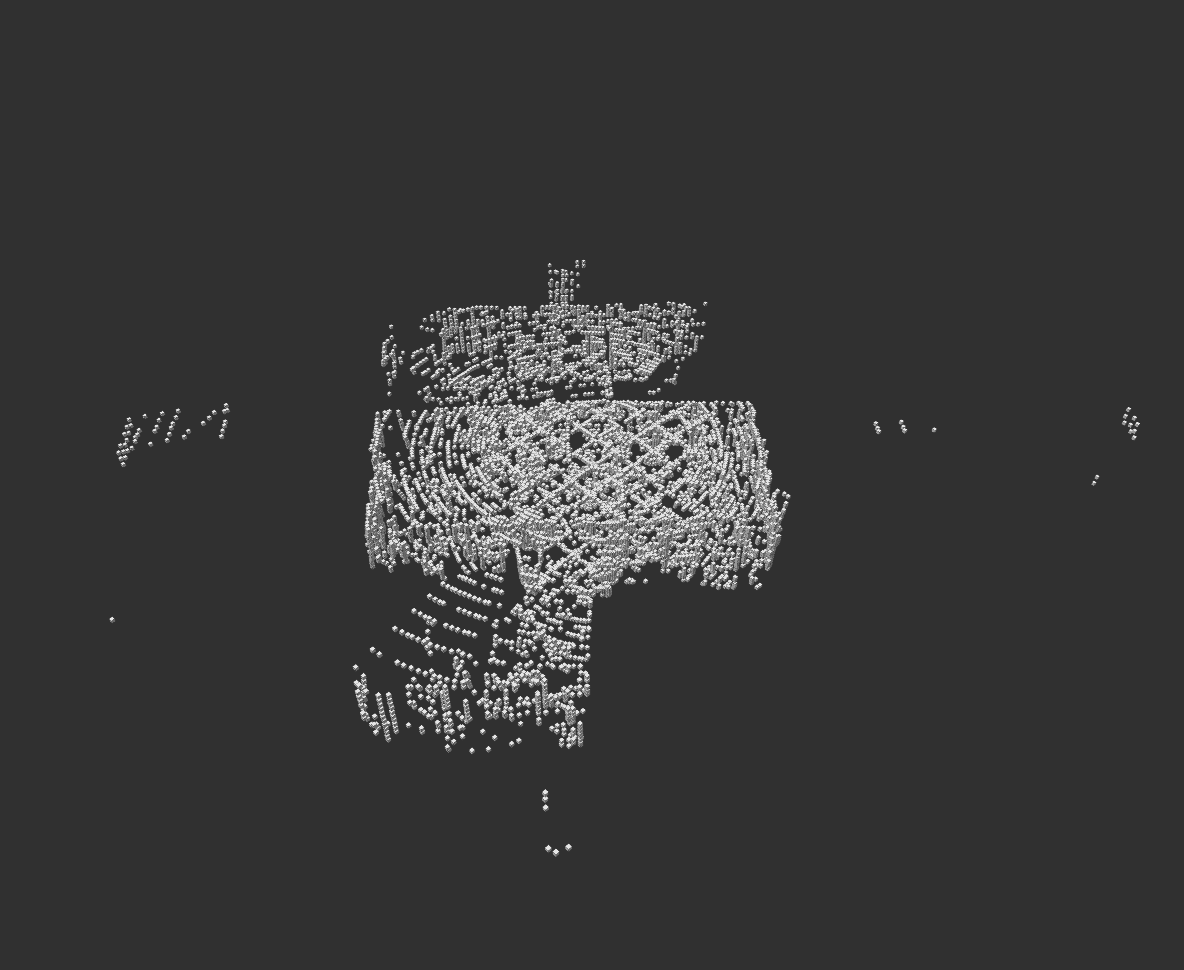}
        \caption{Our method}
        \label{fig:our_method_1}
    \end{subfigure}
    \caption{As shown in the three subfigures above, from left to right: the left subfigure is the original point cloud map, the middle subfigure is the point cloud map processed with a voxel filter, and the right subfigure is our method. Our method preserve critical geometric features.}
    \label{fig:downsampling_illustration}
\end{figure*}

We visualize the result of our mesh reconstruction process after downsampling in Fig.~\ref{fig:mesh_triangulation}. The figure shows that our method successfully retains surface geometry while reducing point cloud density. Table~\ref{tab:runtime_retention} presents the runtime and retention performance at the $7^{\text{th}}$ depth of the tree. We observe that as the point cloud size increases, the retention rate (i.e., the proportion of retained points) decreases. We also compare our method with the commonly used voxel grid filter~\cite{rusu20113d}, which retains a similar percentage of points ($10\%$), as shown in Fig.~\ref{fig:downsampling_illustration}. However, our method selects only the extremal points of the convex hull in each leaf node, enabling better preservation of geometric features while discarding redundant interior points.

\begin{table}[H]
    \centering
    \begin{tabular}{@{}ccc@{}}
        \toprule
        \textbf{Point Cloud Size(k) } & \textbf{Run Time (ms)} & \textbf{Retention Rate (\%)} \\   \midrule
        100  & 41.779   & 79.23 \\ 
        200  & 70.806   & 58.98 \\ 
        300  & 101.226  & 48.34 \\ 
        400  & 115.708  & 41.48 \\ 
        500  & 138.764  & 36.85 \\ 
        600  & 162.865  & 34.03 \\ 
        700  & 181.559  & 31.86 \\ 
        800  & 195.756  & 29.21 \\ 
        900  & 211.645  & 27.30 \\ 
        \bottomrule
    \end{tabular}
 \caption{Run time and retention rate with downsampling.}
    \label{tab:runtime_retention}
\end{table}

\begin{table*}[t]
\centering
\begin{tabularx}{\textwidth}{@{}l*{4}{>{\centering\arraybackslash}X}@{}}
\toprule
\textbf{Cell Size (meter)} & \textbf{Grid Map Dimension} & \textbf{Fixed Uniform Grid} & \textbf{Adaptive Mapping Framework (our method)} & \textbf{Planning Time} \\ \midrule
2.6                & 77$\times$59                       &  683                   & 739            & 401.63                \\ 
2.8                & 73$\times$55                       & 643                  &  721                & 351.52                \\ 
3.0                & 67$\times$51                       &  343                  &  441               & 301.39                \\ 
3.2                & 63$\times$47                       & 177                    & 301                & 238.8                 \\ 
3.4                & 59$\times$45                       &     68                & 128               & 205.0                 \\ \bottomrule
\end{tabularx}
\caption{JPS feasible solution discovery in 1000 trials and run time (microsecond).}
\label{table:times}
\end{table*}

\subsubsection{Jump Point Search}

\begin{figure}[H]
\centering
% 第一行图片
\begin{subfigure}{.23\textwidth}
  \centering
  \includegraphics[width=\linewidth]{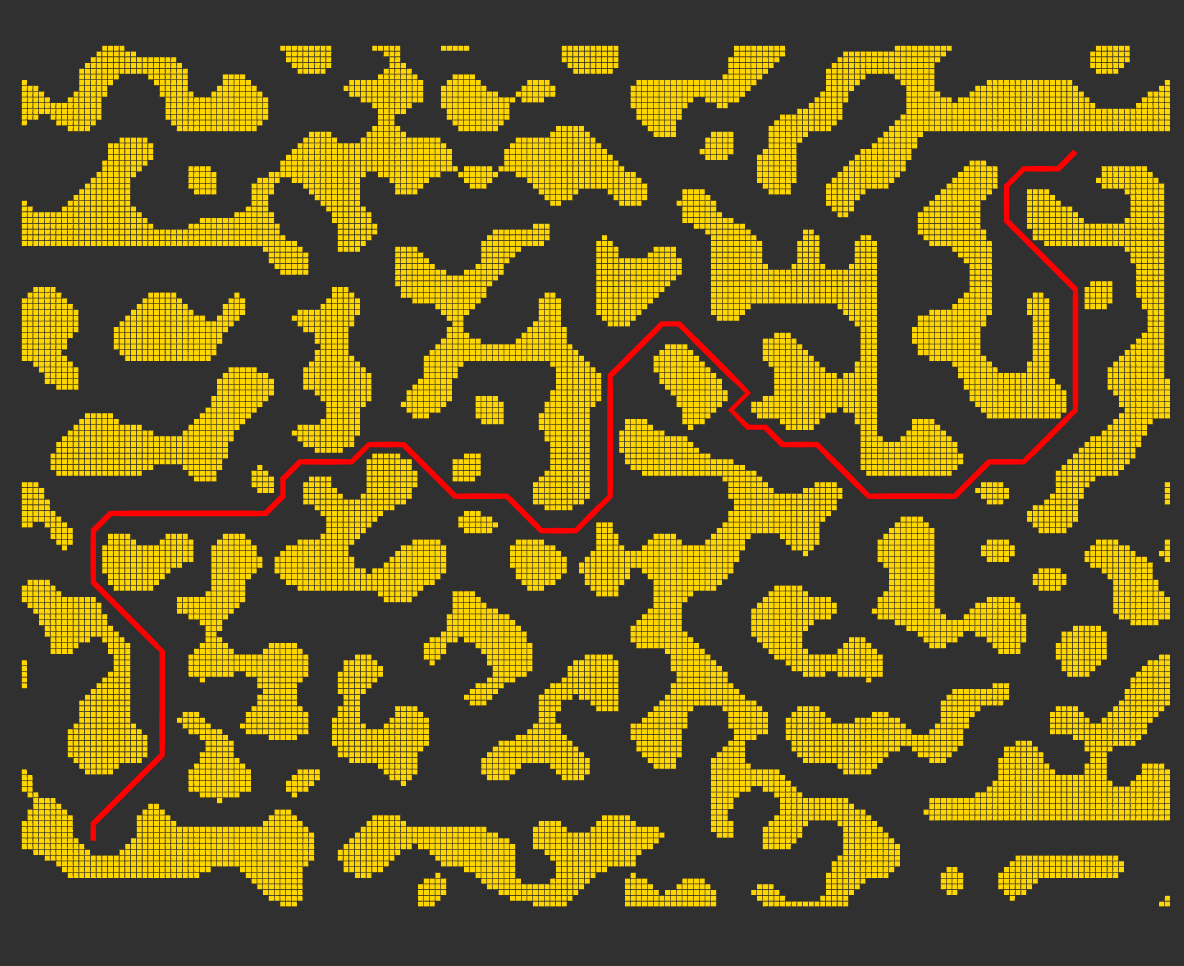}
  % \caption{A}
\end{subfigure}
\begin{subfigure}{.23\textwidth}
  \centering
  \includegraphics[width=\linewidth]{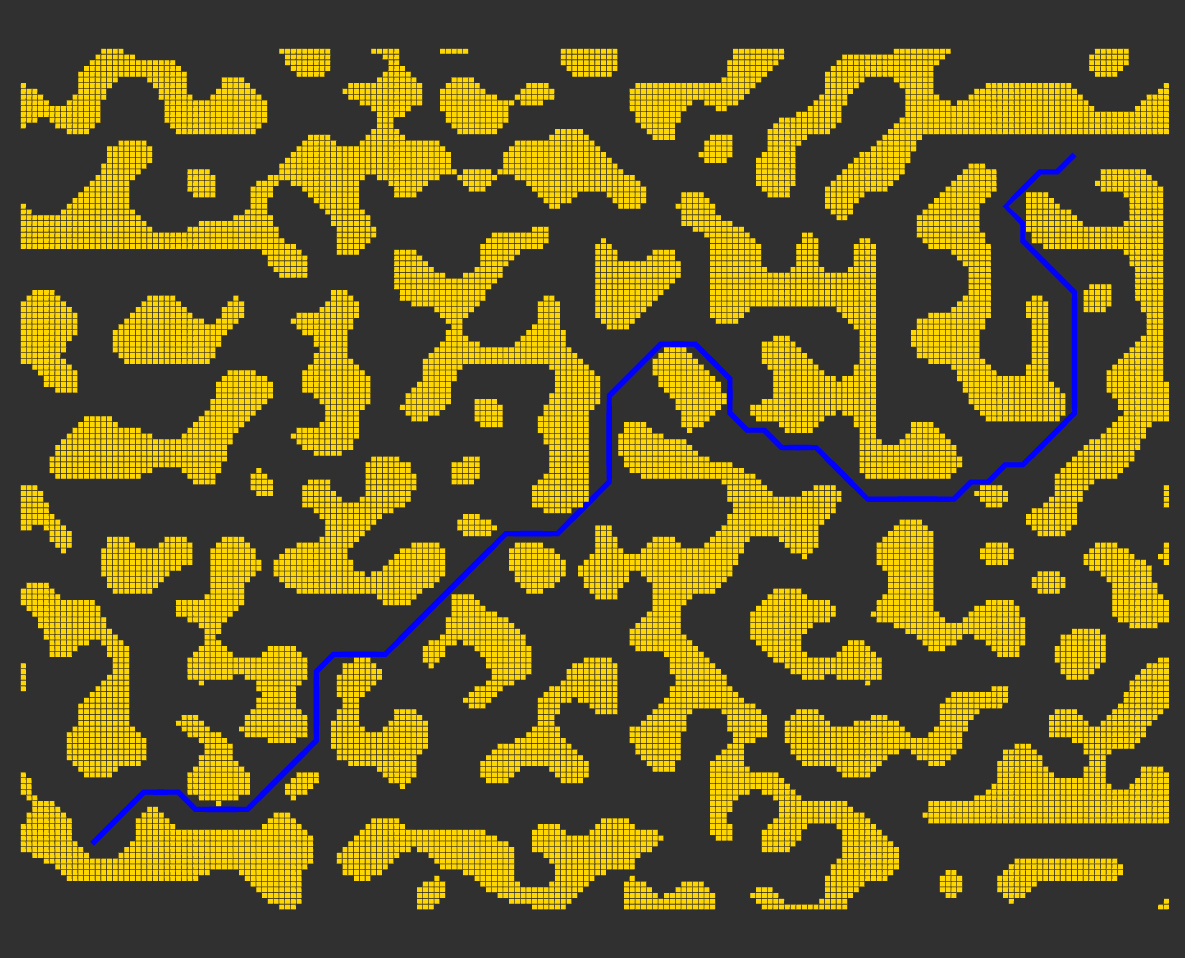}
  % \caption{B}
\end{subfigure}

\caption{Illustration of JPS in the map, with dimensions of $200 \times 150$ meters and approximately 1,200,000 points. The left subfigure (red line path) shows the path finding in the uniform gird map, the right subfigure (blue line path) shows the path finding in our adaptive grid mapping
framework. The path finding using our method is with shorter path length.}
\label{fig:jump point search}
\end{figure}

Figure \ref{fig:jump point search} shows that in the same map, JPS generates paths of different lengths in various types of grid maps. Additionally, in Table \ref{table:times}, we tested 1000 different maps generated by Perlin noise. From both Fig. \ref{fig:jump point search} and Table \ref{table:times}, we can observe that our method has a high probability of finding paths in narrow space environments with shorter path lengths. In all 1000 trials, there was no instance where pathfinding succeeded in the uniform grid but failed with our method.

\begin{figure}[H]
    \centering
    \includegraphics[width=0.50\textwidth]{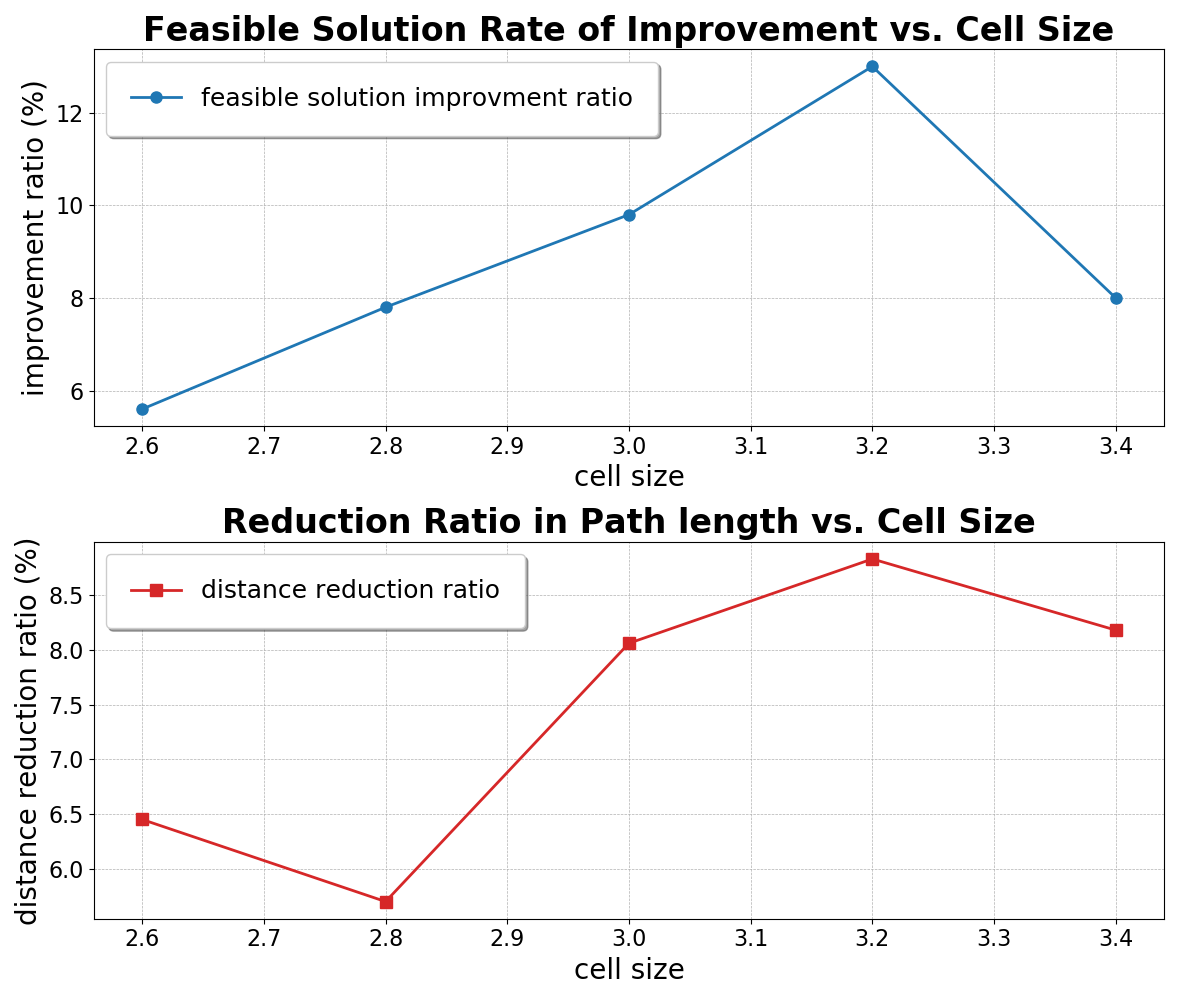}
    \caption{Comparison of feasible path discovery rate and path length reduction, showing that our adaptive mapping consistently improves performance, even in cluttered environments.}
    \label{fig:matric_ratio}
\end{figure}

For rigor, we used Eqn. \eqref{eq:depth} to calculate the tree depth, which allowed us to determine the required number of grids and the grid cell size for my method. We then used a uniform grid of the same cell size for comparison, thus eliminating any size-related advantages. We compared the success rates of feasible solution finding and, in cases where paths were found, the path lengths for both methods across different cell sizes. From the Fig. \ref{fig:matric_ratio}, it is evident that in 1000 maps with dimensions of 200 $\times$ 150 meters and approximately 1,200,000 points, there is an overall improvement of at least $6\%$ in the success rate of finding solutions. Additionally, the path length was reduced by more than $6\%$,  and the run time is less than 0.4 ms.

\section{Conclusion and Future Work}
\label{sec:Conclusion and Future Work}
This paper presents an adaptive grid mapping framework built upon an enhanced OctoMap structure to support efficient and accurate path planning in complex environments. The proposed method introduces a novel tree-based data representation that enables fine-grained segmentation, dynamic spatial refinement, and geometry-aware downsampling. These features collectively facilitate the generation of planning-friendly maps that balance resolution and efficiency. Comprehensive experiments in both simulated and real-world point cloud scenarios demonstrate that our framework achieves higher feasible path discovery rates and produces shorter trajectories compared to traditional uniform-grid-based methods. The framework also supports dynamic updates without full remapping, improving robustness in evolving environments. Future work will explore integration with full planning and control pipelines, scalability to large-scale outdoor scenes, and tighter coupling with real-time sensor data for online operation.
\bibliographystyle{IEEEtran}
\balance
\bibliography{references.bib}
\end{document}